\documentclass[a4paper,twoside]{article}
\usepackage{epsfig}
\usepackage{subcaption}
\usepackage{calc}
\usepackage{amssymb}
\usepackage{amstext}
\usepackage{amsmath}
\usepackage{amsthm}
\usepackage{multicol}
\usepackage{pslatex}
\usepackage{apalike}
\usepackage{algorithm2e}
\usepackage[bottom]{footmisc}
\usepackage{booktabs,array}
\usepackage{tabularx, multirow}
\usepackage[table]{xcolor}
\usepackage{siunitx, url}
\definecolor{kgblue}{HTML}{e0fbfc}
\usepackage{SCITEPRESS}     

\begin{document}
\title{Lie to Me: Knowledge Graphs for Robust Hallucination Self-Detection in LLMs}

\author{\authorname{Sahil Kale\sup{*, 1 }\orcidAuthor{0009-0009-0028-4780} and Antonio Luca Alfeo\sup{*,2 }\orcidAuthor{0000-0002-0928-3188}}
\affiliation{\sup{*}These authors contributed equally to this work}
\affiliation{\sup{1}KnowledgeVerse AI, Arlington, VA, USA}
\affiliation{\sup{2}Dept. of Theoretical and Applied Sciences, eCampus University, Novedrate, Italy}
\email{sahil@k-v.ai, antonioluca.alfeo@uniecampus.it}
}


\abstract{Hallucinations, the generation of apparently convincing yet false statements, remain a major barrier to the safe deployment of LLMs. Building on the strong performance of self-detection methods, we examine the use of structured knowledge representations, namely knowledge graphs, to improve hallucination self-detection. Specifically, we propose a simple yet powerful approach that enriches hallucination self-detection by (i) converting LLM responses into knowledge graphs of entities and relations, and (ii) using these graphs to estimate the likelihood that a response contains hallucinations. We evaluate the proposed approach using two widely used LLMs, GPT-4o and Gemini-2.5-Flash, across two hallucination detection datasets. To support more reliable future benchmarking, one of these datasets has been manually curated and enhanced, and will be released as a secondary outcome of this work. Compared to standard self-detection methods and SelfCheckGPT, a state-of-the-art approach, our method achieves up to 16\% relative improvement in accuracy and 20\% in F1-score. Our results show that LLMs can better analyse atomic facts when they are structured as knowledge graphs, even when initial outputs contain inaccuracies. This low-cost, model-agnostic approach paves the way toward safer and more trustworthy language models. Our dataset and code are publicly available. \footnote{\url{https://github.com/knowledge-verse-ai/kg-hallu-eval}}}

\onecolumn \maketitle \normalsize \setcounter{footnote}{0} \vfill

\section{\uppercase{Introduction}}
\label{sec:introduction}
Among the many challenges facing Large Language Models (LLMs), their tendency to hallucinate stands as the single most formidable obstacle to responsible deployment in critical applications \cite{alfeo2021technological}. Hallucination detection in LLMs seeks to identify instances where LLMs generate either uncertain outputs or statements that appear plausible but are factually incorrect \cite{jiang-etal-2024-large}. This issue has been described both as a mysterious, persistent challenge \cite{xu2025hallucinationinevitableinnatelimitation} and as a straightforward error in binary classification \cite{kalai2025languagemodelshallucinate}.

Whatever the reason, the undeniable fact is that hallucinations severely undermine the trustworthiness and practical utility of language models. While elimination of hallucinations still requires a significant amount of work, precise and simple detection can greatly help trigger correction mechanisms and reduce the spread of misinformation. 

Self-detection methods \cite{11071605}, or methods where the generator and detector of the hallucination are identical \cite{valentin2024costeffectivehallucinationdetectionllms}, offer a promising branch of flagging hallucinations in LLMs. They offer key advantages, notably lower cost and the elimination of external dependencies. Most current self-detection techniques employ reprocessing a single sample to check its faithfulness \cite{manakul-etal-2023-selfcheckgpt}, or aggregate internal confidence and consistency from multiple samples \cite{ji-etal-2023-towards}. However, we observe that most existing methods lack structured representations during reprocessing and aggregation. We show that structuring model representations with knowledge graphs, which organise facts as interconnected entities and relations \cite{agrawal-etal-2024-knowledge}, activates latent reasoning and reflection capabilities, leading to substantial improvements in hallucination self-detection performance.

\begin{figure}[t]
    \centering
    \begin{subfigure}[b]{0.46\columnwidth}
        \centering
        \includegraphics[height=2.5cm, keepaspectratio]{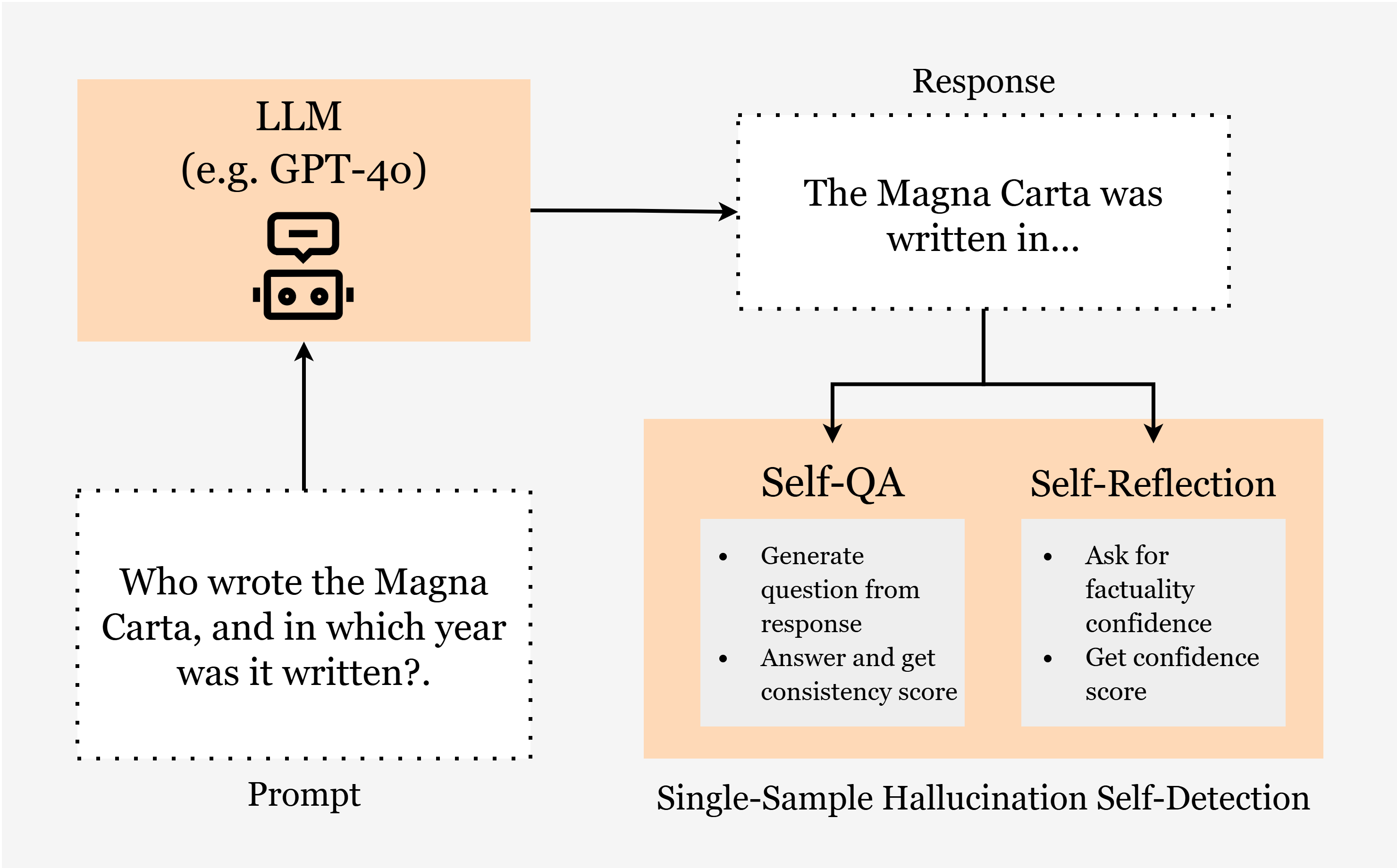}
        \caption{Standard self-detection flow}
        \label{fig:kg1a}
    \end{subfigure}
    \hfill
    \begin{subfigure}[b]{0.46\columnwidth}
        \centering
        \includegraphics[height=2.5cm, keepaspectratio]{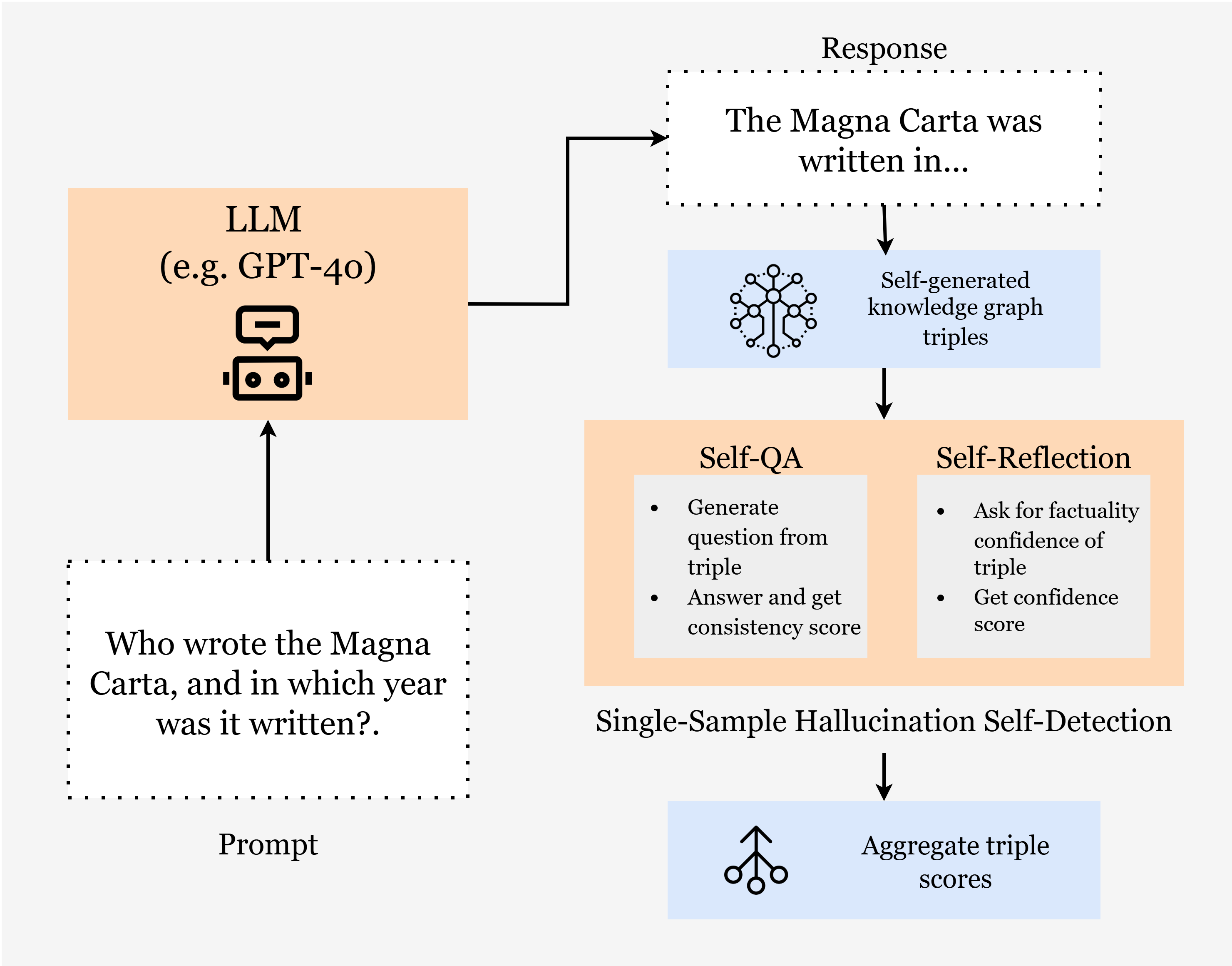}
        \caption{Self-detection flow with knowledge graph}
        \label{fig:kg1b}
    \end{subfigure}
    \vspace{0.2cm}
    \caption{Single-sample hallucination self-detection flow for LLMs and the proposed knowledge graph augmented method.}
    \label{fig:kg1}
\end{figure}

Our motivation addresses a central limitation in hallucination self-detection. Prior work shows that introducing structure into model reasoning improves analysis of constituent facts \cite{agrawal-etal-2024-knowledge}. However, existing single-sample self-detection methods typically treat generated responses as undifferentiated text, forcing confidence or consistency to be assessed globally and obscuring which specific claims are reliable or hallucinated. Knowledge graphs (KGs) provide structured representations in which entities are nodes and relations are labelled edges. By decomposing responses into discrete, connected facts, our approach enables targeted and transparent analysis. Moreover, in multi-sample settings, similarity-based aggregation can be misleading, as identifying precise semantic agreement among partially conflicting responses is inherently difficult \cite{valentin2024costeffectivehallucinationdetectionllms}. Representing knowledge in a fixed, structured form reduces this problem to direct comparisons of entities and relations, improving both reliability and interpretability.
The key contributions from our work can be summarised as follows:
\begin{enumerate}
    \item We show that introducing knowledge graphs can significantly enhance current single and multi-sample techniques, boosting F1 scores by up to 20.4\% and AUC-PR by up to 21.0\% relative improvement for popular models like GPT-4o and Gemini 2.5 Flash.
    \item We release a manually curated dataset targeted for hallucination self-detection, covering over 500 examples with realistic, controlled hallucinations and actual model-generated samples.
\end{enumerate}

\section{\uppercase{Related Work}}
\label{sec:related-work}
\subsection{Hallucination Self-Detection in LLMs}
Hallucination detection aims to distinguish hallucinated from factual content in LLM outputs \cite{hallu-defn}. Existing approaches are commonly categorised by their objective or technical design \cite{hallu-survey,siren}. By objective, methods target either factuality detection \cite{wang-etal-2024-factuality}, which identifies objective inaccuracies using external knowledge, or faithfulness detection, which evaluates alignment with the given context and the model’s internal knowledge, often through confidence-based analysis \cite{sahoo-etal-2024-comprehensive}.

From a technical perspective, detection methods include natural language inference pipelines such as FACTOID \cite{rawte2024factoidfactualentailmenthallucination}. A prominent subclass of these approaches is self-detection, which relies solely on a model’s internal mechanisms and requires no external knowledge or resources \cite{ji-etal-2023-towards}. This makes self-detection simple to deploy and broadly applicable, particularly in settings with limited access to verified data sources \cite{sahoo-etal-2024-comprehensive}.

Early self-detection methods focused on multi-sample techniques that aggregate confidence and consistency across multiple regenerations, with representative examples including SelfCheckGPT \cite{manakul-etal-2023-selfcheckgpt} and Demonstrative Exemplars \cite{varshney2024investigatingaddressinghallucinationsllms}. While their minimal external dependencies and zero-shot black box applicability make them attractive for general-purpose use, repeated LLM calls incur substantial computational cost \cite{hallu-survey}. In contrast, single-sample self-detection methods operate in a single inference pass through structured reasoning or analysis, as exemplified by Self-Contradictory Reasoning \cite{liu-etal-2024-self-contradictory} and verbalised confidence analysis \cite{kale-vrn-2025-line}. These approaches are more efficient and cost-effective, though generally less robust across prompts, domains, and model behaviours than multi-sample methods \cite{valentin2024costeffectivehallucinationdetectionllms}.

\begin{figure}[t]
    \centering
    \begin{subfigure}[b]{0.46\columnwidth}
        \centering
        \includegraphics[height=2.5cm, keepaspectratio]{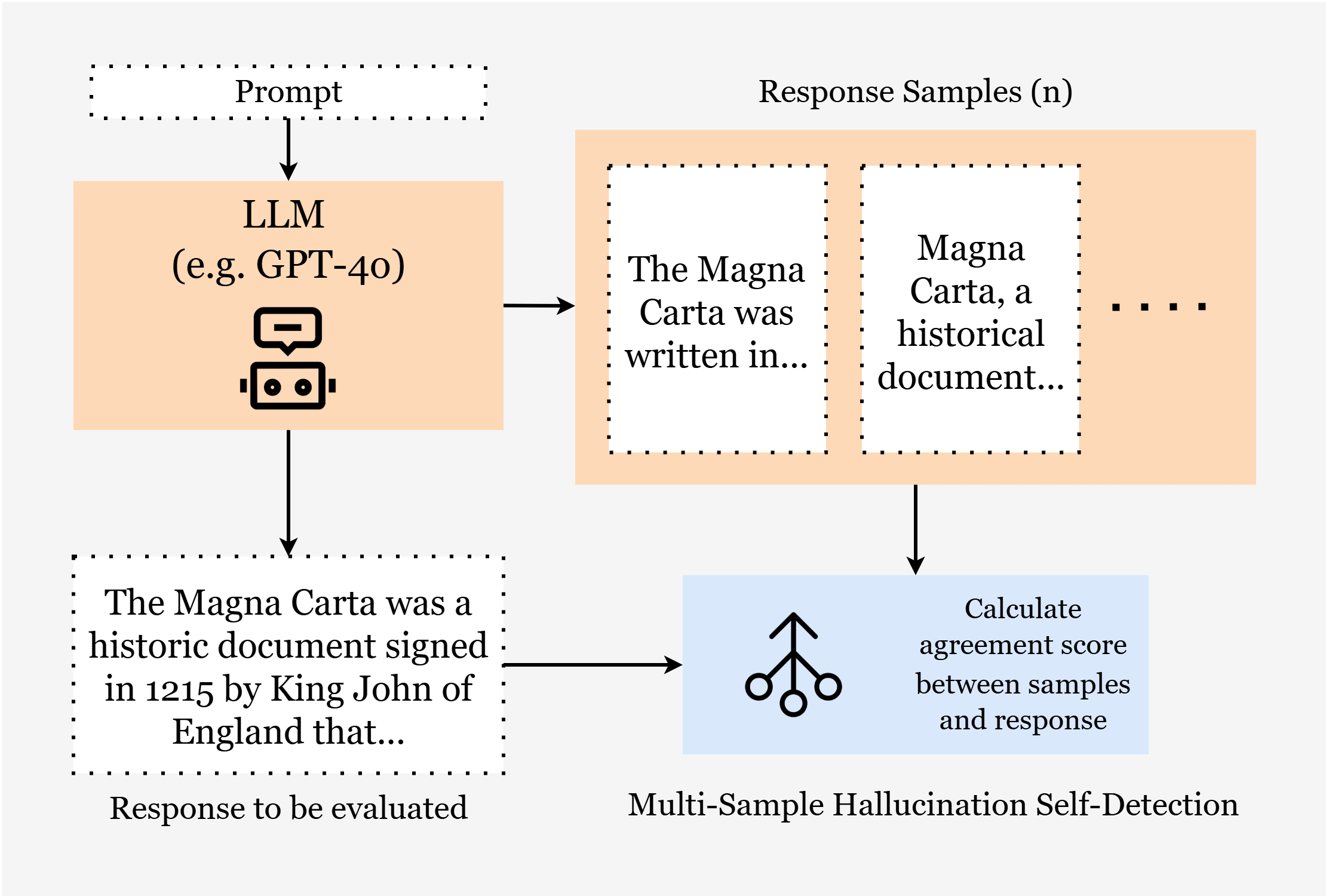}
        \caption{Standard multi-sample flow}
        \label{fig:kg2a}
    \end{subfigure}
    \hfill
    \begin{subfigure}[b]{0.46\columnwidth}
        \centering
        \includegraphics[height=2.5cm, keepaspectratio]{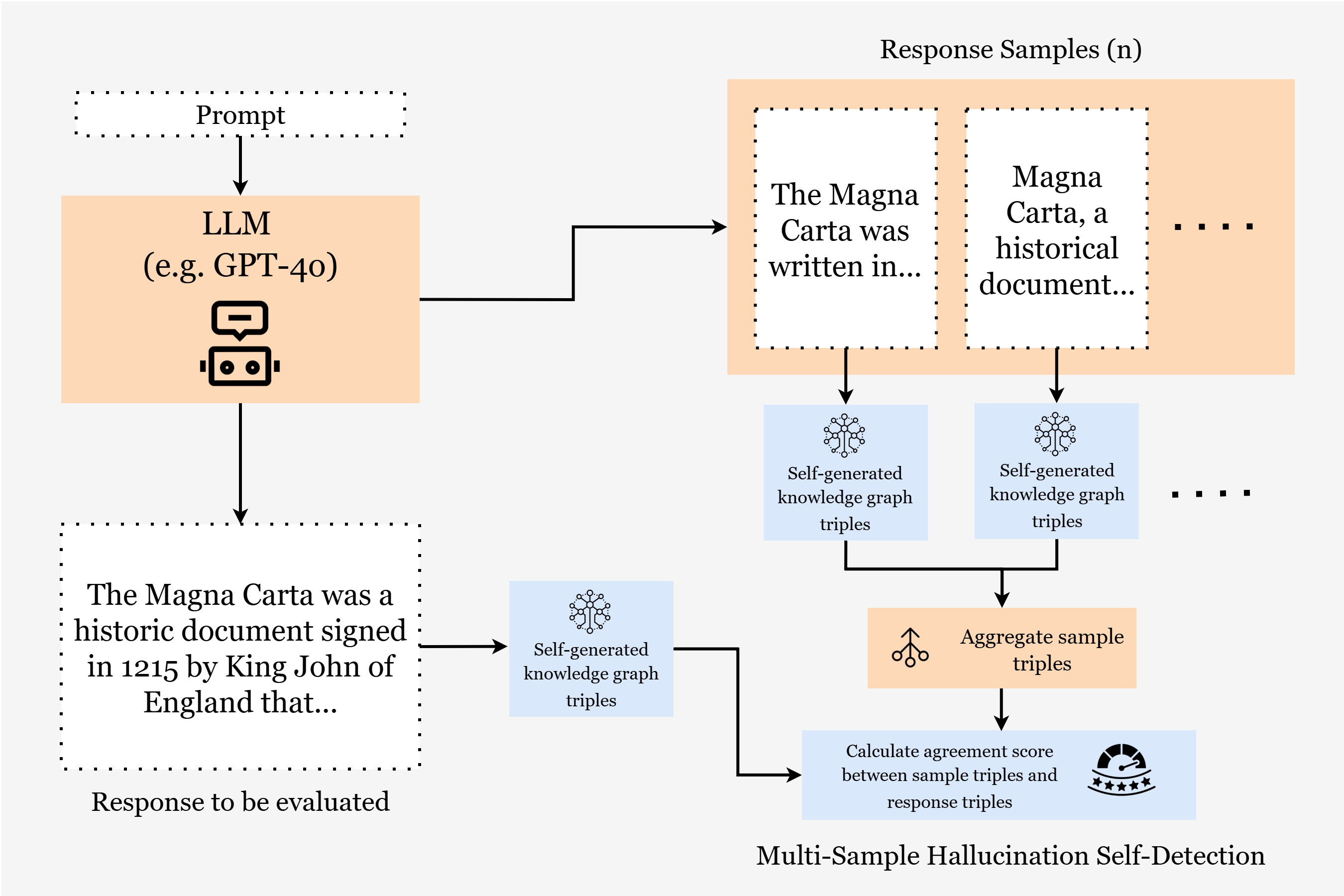}
        \caption{Flow with knowledge graph}
        \label{fig:kg2b}
    \end{subfigure}
    \vspace{0.2cm}
    \caption{Multi sample hallucination self-detection flow for LLMs and the proposed knowledge graph augmented method.}
    \label{fig:kg2}
\end{figure}

\subsection{Knowledge Graphs for Hallucination Detection}
Knowledge graphs provide structured representations that support reasoning and retrieval, and have been widely integrated into LLMs to improve performance. Existing approaches primarily incorporate KG knowledge either during training \cite{wang-train} or at inference time \cite{kim-etal-2023-kg}, including through prompting, architectural adaptations, or post-generation refinement \cite{LAVRINOVICS2025100844}.

Recent work has explored the use of knowledge graphs to mitigate hallucinations \cite{agrawal-etal-2024-knowledge}. Methods such as Think-on-Graph \cite{sun2023thinkongraph} and ChatTf \cite{chat-tf} demonstrate that graph-based reasoning and traversal can structure contextual relations and improve factual consistency.

\section{\uppercase{Methodology}}
In this section, we elaborate on how incorporating self-generated knowledge graphs in the hallucination self-detection flow of existing methods can improve performance and enable more accurate identification. For simplicity, we assume that all responses to be analysed for hallucination are in a single sentence, and possibly in the context of a paragraph.

\subsection{Knowledge Graph Construction}
\label{sec:kg-build}
Owing to the strong capabilities of recent LLMs, complex multi-step pipelines previously used for entity and relation extraction \cite{zhang-soh-2024-edc} are no longer necessary. Instead, a single, carefully designed prompt guides the model to extract a structured set of factual triples in the form of (subject, relation, object) from each output. These triples are then assembled into a knowledge graph that can be integrated into the information flow of hallucination self-detection techniques. 

Formally, given an LLM generated output \( O \), the constructed knowledge graph is defined as
\begin{equation}
\text{KG}_O = \{ (e_i, r_i, e_j) \mid e_i, e_j \in E, \ r_i \in R \},
\end{equation}
where \( E \) is the set of extracted entities and \( R \) is the set of labelled relations between them. For simplicity, each triple is denoted as \( t \in \text{KG}_O \).

\subsection{Single-Sample Methods}
With Single-Sample methods, an LLM is asked to analyse or reprocess a response for a given prompt only once. Essentially, they are based on detecting and scoring a hallucination from a single generated output. Each method produces a \textit{score} between 0 and 1, where lower values indicate higher likelihood of hallucination. While the computation of this score differs by technique, ranging from probabilistic estimates to direct LLM scoring, it is standard practice in methods that interpret hallucination detection as binary classification problems \cite{hallu-survey}. We analyse and improve two single-sample methods applicable for both open-source and black-box models. 

\vspace{0.1cm}
\noindent\textbf{Self-Questioning:} We emulate the Chain-of-Thought verification and slightly modify it for the hallucination self-detection objective. CoVe (Chain-of-Verification) \cite{dhuliawala-etal-2024-chain} shows that having a model draft questions about its own answer and then answer them reduces hallucinations. We adapt this by asking the LLM to self-interrogate each response it generates and score consistency such that a lower consistency score indicates possible hallucinations. 

In the Self-Questioning method, the LLM \( P \) performs internal consistency verification through a three-step process applied to the output \( o_i \), as shown in Figure \ref{fig:kg1a}:

\begin{enumerate}
    \item Given an output \( o_i \), the model is first prompted to generate a verification question \( q_i \) that probes its factual soundness.  
    \item The model is then asked to answer this question, yielding a self-generated response \( a_i \).  
    \item Finally, the model is prompted to evaluate the agreement between the original output and its own answer, producing a consistency score \( C_i \in \{0,1\} \). Here, \( P_{\text{consistency}} \) denotes the score provided by the model when asked to estimate the consistency and agreement between \( o_i \) and \( a_i \).
\end{enumerate}
\begin{equation}
    C_i = P_{\text{consistency}}(o_i, a_i)
    \label{eq:selfq-consistency}
\end{equation}

The final score \( C_i \) indicates whether the original output \( o_i \) is considered factual (closer to 1) or potentially hallucinated (closer to 0). 

\vspace{0.1cm}
\noindent\textbf{Self-Questioning with Knowledge Graphs:} While the text-based Self-Questioning method is effective, its main limitation lies in analysing the entire generated response \( o_i \) as a single block. A single sentence often encodes multiple facts, with number and complexity varying across contexts and domains. Without treating outputs as structured collections of related facts, LLMs struggle to precisely localise and score hallucinated content \cite{xue2025verify}.

To address this issue, we construct a knowledge graph from the generated response (using the procedure in Section \ref{sec:kg-build}), which allows us to analyse hallucinations at the level of individual facts rather than large text. Each fact in the response is represented as a triple \( t_j \in KG_{o_i} \). For each triple, the model is prompted to generate a verification question, provide an answer, and then assign a consistency score. This yields a consistency score \( c_j \in [0,1] \) for each triple, defined as: 
\begin{equation}
c_j = P_{\text{consistency}}(t_j, a_i)
\label{eq:triple_score}
\end{equation}
where \( P_{\text{consistency}} \) denotes the score provided by the model when asked to estimate the consistency and agreement between the triple \( t_j \) and its corresponding answer \( a_i \).  

Finally, the overall score for the output \( o_i \) is obtained by averaging the score across all triples in the knowledge graph:  
\begin{equation}
C_i = \frac{1}{|KG_{o_i}|} \sum_{j=1}^{|KG_{o_i}|} c_j
\label{eq:output_score}
\end{equation}

\vspace{0.1cm}
\noindent\textbf{Self-Confidence:} Another single-sample approach for hallucination self-detection is to directly elicit confidence scores from the model itself, inspired by recent research \cite{lin2022teachingmodelsexpressuncertainty}. Confidence elicitation provides a direct, interpretable signal of the model’s belief in its own outputs, without requiring intermediate question–answer steps. 

In practice, to make the method applicable for black-box settings, the LLM is prompted to output a verbatim confidence score in the range \([0,1]\) accompanying the response \( o_i \). A lower confidence score is taken to imply lower factuality and reflect greater uncertainty and potential hallucination. In the Self-Confidence method, the LLM performs confidence elicitation through the following process applied to the output \( o_i \), as shown in Figure \ref{fig:kg1a}:

\begin{enumerate}
    \item The model provides a response \( o_i \) to a given input prompt \( i\).
    \item The model is explicitly prompted to provide a self-assessed confidence score \( C_i \in [0,1] \) for the factual accuracy of the response \( o_i \).
\end{enumerate}

Formally, the confidence score is represented as:
\begin{equation}
C_i = P_{\text{confidence}}(o_i)
\end{equation}

\vspace{0.1cm}
\noindent\textbf{Self-Confidence with Knowledge Graphs:} As with Self-Questioning, we argue that the main limitation of sentence-level confidence elicitation is treating the entire output \( o_i \) as a single block. By splitting the response into a knowledge graph, we provide a structured representation for the model to reflect on.

Toward this objective, we represent each fact as a knowledge graph triple \( t_j \in KG_o \), and the model is prompted to provide a confidence score \( c_j \in [0,1] \) for each fact individually, as shown in Figure \ref{fig:kg1b}. The final confidence score \( C_i \) for the response is then computed as the average of all triple-level confidences:
\begin{equation}
c_j = P_{\text{confidence}}(t_j)
\end{equation}
\begin{equation}
C_i = \frac{1}{|KG_{o_i}|} \sum_{j=1}^{|KG_{o_i}|} c_j
\label{eq:output_score}
\end{equation}

\subsection{Multi-Sample Methods}
Multi-sample methods detect hallucination by generating multiple responses from the same LLM given an identical input, and comparing consistency with the response under evaluation. The core principle is that an LLM is less likely to hallucinate the same false content consistently across multiple generations \cite{manakul-etal-2023-selfcheckgpt,xue2025verify}. 

Multi-sample methods provide a score between 0 and 1 to show the probability of a response being accurate. We show improvement in the SelfCheckGPT \cite{manakul-etal-2023-selfcheckgpt} approach by using knowledge graphs in the flow. Other multi-sample methods, like NeMO Guardrails \cite{dong2024buildingguardrailslargelanguage}, are incompatible with our approach because they require concatenating multiple samples into a single input, exceeding the LLM’s context length.

\vspace{0.1cm}
\noindent\textbf{SelfCheckGPT:} SelfCheckGPT is a black-box, zero-resource hallucination detection method that measures the consistency of an LLM’s output by comparing it against multiple stochastic samples generated for the same prompt. For a language model \( P \) has produced an output sample \( o_i \) for a given prompt \( i \). We generate \( n \) additional output samples \( \{s_1, s_2, \dots, s_n\} \) using the same prompt. The BERTScore \( C_i \) (named from the original reference \cite{manakul-etal-2023-selfcheckgpt}) for the output \( o_i \) is computed as the average semantic similarity between \( o_i \) and the samples, where a higher score indicates greater agreement and hence higher confidence in factuality:

\begin{equation}
C_i = \frac{1}{n} \sum_{j=1}^{n} \text{sim}(o_i, s_j)
\label{eq:selfcheckgpt-sim}
\end{equation}
where \( \text{sim}(\cdot, \cdot) \) denotes any semantic similarity function. In our case, we use the SBERT embeddings \cite{all-MiniLM-L6-v2} to compute similarity.

\vspace{0.1cm}
\noindent\textbf{SelfCheckGPT with Knowledge Graphs:}  
We enhance SelfCheckGPT by incorporating knowledge graphs to perform fact-level consistency analysis. Assume that a language model \( P \) has produced an output sample \( o_i \) for a given prompt \( i \), and \( n \) additional output samples \( \{s_1, s_2, \dots, s_n\} \) are generated using the same prompt.  

Both the original response \( o_i \) and all \( n \) sampled responses \( \{s_1, s_2, \dots, s_n\} \) are converted into knowledge graphs using the process described in Section \ref{sec:kg-build}. For each triple \( t_j \in KG_{o_i} \), we compute the average semantic similarity between \( t_j \) and the most similar triple in each sample graph \( KG_{s_k} \), where \( k \in \{1, \dots, n\} \). The similarity is computed computed using the cosine distance between the embeddings obtained by processing the triplets with SBERT \cite{all-MiniLM-L6-v2}. The per-triple BERTScore score \( c_j \) is then averaged over all triples in the original graph to obtain the final BERTScore \( C_i \) for the original response.

\begin{equation}
c_j = \frac{1}{n} \sum_{k=1}^{n} \max_{t' \in KG_{s_k}} \text{sim}(t_j, t')
\label{eq:triple-score}
\end{equation}

\begin{equation}
C_i = \frac{1}{|KG_{o_i}|} \sum_{j=1}^{|KG_{o_i}|} c_j
\label{eq:final-score}
\end{equation}

\section{\uppercase{Experimental Setup}}
\subsection{Model Setup}
To ensure comprehensive analysis and a true reflection of real-world performance, we experiment with two high-performance models. Since our methodology is mainly intended to improve black-box models, we choose two closed-source models, GPT-4o \cite{openai2025gpt4o} and Gemini 2.5 Flash \cite{comanici2025gemini25pushingfrontier} in our evaluation. All models were accessed through provider APIs, and common parameter values set during knowledge graph generation (more deterministic) and single-sample flows (less deterministic) are listed in Table \ref{tab:merged-params}. We run our experiments with GPT-4o by accessing it via the OpenAI Python SDK. We also use the recently released model, Gemini 2.5 Flash, through the Google Generative AI Python SDK via API calls.

\renewcommand{\arraystretch}{1.05}
\begin{table}[t]
\centering
\caption{Model parameters used for KG construction and self-detection techniques}
\small
\begin{tabular}{lrr}
\toprule
\textbf{Parameter} & \textbf{KG} & \textbf{Self-detection} \\
\midrule
\texttt{temperature}        & 0.0  & 1.0 \\
\texttt{top\_p}             & 1.0  & 1.0 \\
\texttt{max\_tokens}        & 8096 & 8096 \\
\texttt{frequency\_penalty} & 1.0  & 0.0 \\
\texttt{presence\_penalty}  & 1.0  & 0.0 \\
\bottomrule
\end{tabular}
\label{tab:merged-params}
\end{table}

\subsection{Metrics and Evaluation}
Reliable hallucination detection systems must provide both calibrated confidence scores and effective binary decisions on whether an output is hallucinated. However, because both the scoring prompts and the way models interpret those prompts vary across methods, the computed scores are not directly comparable at every threshold. Hence, we perform a simple linear hyper-parameter search to identify the optimal threshold (maximising either accuracy or F1 score independently) for each method, i.e. a cut-off point above which an output is classified as factual and below which it is classified as hallucinated. We report these best-case binary classification results in Table \ref{tab:gpt-gemini-merged} alongside the global AUC-PR score, which remains threshold-independent. For multi-sample methods, we use a fixed set of $n=20$ samples. 

\subsection{Datasets}
We conduct experiments on two primary datasets. For multi-sample methods, due to the higher computational cost of generating multiple responses per query, we focus our evaluation on only one dataset.

\noindent\textbf{SimpleQA:} SimpleQA \footnote{\url{https://huggingface.co/datasets/basicv8vc/SimpleQA}} \cite{simpleqa} is a factuality benchmark provided by OpenAI that evaluates a language model’s ability to answer short fact-seeking questions. It contains 4.3k question-answer pairs from various domains. Following the evaluation protocol provided in the benchmark, we first classify samples as accurate or inaccurate. Because SimpleQA had few correct model answers, we balanced the dataset by selecting equal numbers of correct and incorrect samples (total 1550 sentences for GPT-4o, 990 for Gemini-2.5-Flash). 

\noindent\textbf{WikiBio GPT-4o:} The second dataset is our manually curated WikiBio GPT-4o \footnote{\url{https://github.com/knowledge-verse-ai/kg-hallu-eval}} hallucination detection benchmark, created to address key limitations in existing resources. Most current datasets either lack real-world LLM-generated samples or suffer from class imbalance, as seen in WikiBio GPT-3 \cite{manakul-etal-2023-selfcheckgpt}, which relies on outdated models and over-represents hallucinated examples. Our dataset contains 501 carefully curated sentences, sampled from real outputs of the modern GPT-4o model, and is balanced and high-quality, supporting robust evaluation of both single-sample and multi-sample hallucination detection methods. 

\subsubsection{WikiBio GPT-4o Construction}
Table \ref{tab:data-stats} summarises the dataset statistics in comparison with WikiBio GPT-3, including class counts and length statistics for accurate and hallucinated sentences. While smaller in scale, the WikiBio GPT-4o dataset exhibits several advantages: a higher average number of sentences per paragraph, more balanced distributions of accurate and hallucinated sentences, and slightly higher semantic similarity with comparable sentence lengths across classes. These characteristics make it better suited for evaluating nuanced hallucination detection. In addition, the longer average sample length enables models to capture richer contextual information, improving the reliability of downstream evaluations.

The dataset construction closely follows the WikiBio GPT-3 methodology. Introductory paragraphs of 50 Wikipedia articles, each approximately 200 words, were randomly selected as base texts. Using GPT-4o \cite{openai2025gpt4o} with a temperature of 1.0, 20 samples were generated per article. Original texts were segmented into sentences, after which hallucinated examples were created either by inserting falsified sentences or by modifying existing ones with incorrect information, under strict manual supervision by a STEM graduate-level expert. Each sentence was then annotated as accurate or hallucinated.

\renewcommand{\arraystretch}{1.05}
\begin{table}[t]
\centering
\caption{WikiBio GPT-3 and GPT-4o dataset statistics}
\small
\begin{tabular}{lrr}
\toprule
\textbf{Metric} & \textbf{GPT-3} & \textbf{GPT-4o} \\
\midrule
Sentences (total) & 1,903 & 501 \\
Paragraphs (total) & 238 & 50 \\
Sentences per paragraph & 8 & 10 \\
Hallucinated sentences & 1,392 & 241 \\
Accurate sentences & 516 & 260 \\
Samples per paragraph & 20 & 20 \\
Avg. sample length (words) & 146 & 169 \\
Avg. hallucinated length (words) & 17 & 14 \\
Avg. accurate length (words) & 18 & 14 \\
Semantic similarity (H vs. A) & 0.30 & 0.37 \\
\bottomrule
\end{tabular}
\label{tab:data-stats}
\end{table}

\section{\uppercase{Results and Discussion}}
\renewcommand{\arraystretch}{1.1}
\begin{table*}[t]
\centering
\caption{Performance comparison of hallucination self-detection methods for GPT-4o and Gemini-2.5-Flash models (with and without knowledge graph support, +KG). Values represent the average at the best decision threshold with 95\% confidence intervals (mean $\pm$ 95\% CI).}
\small
\resizebox{\textwidth}{!}{%
\begin{tabular}{%
    >{\centering\arraybackslash}p{0.13\textwidth}  
    >{\centering\arraybackslash}p{0.15\textwidth}  
    >{\centering\arraybackslash}p{0.15\textwidth}  
    >{\centering\arraybackslash}p{0.15\textwidth}  
    >{\centering\arraybackslash}p{0.15\textwidth}  
    >{\centering\arraybackslash}p{0.15\textwidth}  
    >{\centering\arraybackslash}p{0.15\textwidth}  
    >{\centering\arraybackslash}p{0.15\textwidth}  
}
\toprule
\multirow{2}{*}{\textbf{Dataset}} & \multirow{2}{*}{\textbf{Method}} &
\multicolumn{3}{c}{\textbf{GPT-4o}} &
\multicolumn{3}{c}{\textbf{Gemini-2.5-Flash}} \\
\cmidrule(lr){3-5} \cmidrule(lr){6-8}
& & \textbf{Accuracy} & \textbf{F1} & \textbf{AUC-PR} & \textbf{Accuracy} & \textbf{F1} & \textbf{AUC-PR} \\
\midrule
\multicolumn{8}{c}{\textbf{Single-Sample Methods}}\\
\midrule
\multirow{4}{*}{SimpleQA}%
 & Self-Questioning
   & 0.60 $\pm$ 0.029 & 0.51 $\pm$ 0.026 & 0.57 $\pm$ 0.042
   & 0.62 $\pm$ 0.034 & 0.49 $\pm$ 0.043 & 0.50 $\pm$ 0.051 \\
 & \cellcolor{kgblue}\makebox[1.5cm][r]{\textit{+KG}}
   & \cellcolor{kgblue}0.67 $\pm$ 0.027
   & \cellcolor{kgblue}0.60 $\pm$ 0.030
   & \cellcolor{kgblue}0.69 $\pm$ 0.037
   & \cellcolor{kgblue}0.63 $\pm$ 0.034
   & \cellcolor{kgblue}0.59 $\pm$ 0.037
   & \cellcolor{kgblue}0.62 $\pm$ 0.041 \\
 & Self-Confidence
   & 0.57 $\pm$ 0.028 & 0.57 $\pm$ 0.029 & 0.50 $\pm$ 0.042
   & 0.53 $\pm$ 0.033 & 0.57 $\pm$ 0.033 & 0.53 $\pm$ 0.050 \\
 & \cellcolor{kgblue}\makebox[1.5cm][r]{\textit{+KG}}
   & \cellcolor{kgblue}0.64 $\pm$ 0.028
   & \cellcolor{kgblue}0.63 $\pm$ 0.028
   & \cellcolor{kgblue}0.61 $\pm$ 0.042
   & \cellcolor{kgblue}0.56 $\pm$ 0.033
   & \cellcolor{kgblue}0.59 $\pm$ 0.033
   & \cellcolor{kgblue}0.53 $\pm$ 0.053 \\
\midrule
\multirow{4}{*}{WikiBio GPT-4o}%
 & Self-Questioning
   & 0.72 $\pm$ 0.037 & 0.71 $\pm$ 0.044 & 0.78 $\pm$ 0.042
   & 0.62 $\pm$ 0.041 & 0.63 $\pm$ 0.048 & 0.63 $\pm$ 0.057 \\
 & \cellcolor{kgblue}\makebox[1.5cm][r]{\textit{+KG}}
   & \cellcolor{kgblue}0.79 $\pm$ 0.035
   & \cellcolor{kgblue}0.81 $\pm$ 0.034
   & \cellcolor{kgblue}0.84 $\pm$ 0.038
   & \cellcolor{kgblue}0.69 $\pm$ 0.041
   & \cellcolor{kgblue}0.73 $\pm$ 0.042
   & \cellcolor{kgblue}0.73 $\pm$ 0.050 \\
 & Self-Confidence
   & 0.79 $\pm$ 0.034 & 0.81 $\pm$ 0.034 & 0.86 $\pm$ 0.037
   & 0.78 $\pm$ 0.036 & 0.79 $\pm$ 0.040 & 0.83 $\pm$ 0.048 \\
 & \cellcolor{kgblue}\makebox[1.5cm][r]{\textit{+KG}}
   & \cellcolor{kgblue}0.85 $\pm$ 0.028
   & \cellcolor{kgblue}0.87 $\pm$ 0.029
   & \cellcolor{kgblue}0.88 $\pm$ 0.035
   & \cellcolor{kgblue}0.84 $\pm$ 0.031
   & \cellcolor{kgblue}0.85 $\pm$ 0.030
   & \cellcolor{kgblue}0.87 $\pm$ 0.041 \\
\midrule
\multicolumn{8}{c}{\textbf{Multi-Sample Methods}}\\
\midrule
\multirow{2}{*}{WikiBio GPT-4o}%
 & Self-Check GPT
   & 0.63 $\pm$ 0.041 & 0.69 $\pm$ 0.039 & 0.62 $\pm$ 0.069
   & 0.60 $\pm$ 0.042 & 0.68 $\pm$ 0.037 & 0.55 $\pm$ 0.057 \\
 & \cellcolor{kgblue}\makebox[1.5cm][r]{\textit{+KG}}
   & \cellcolor{kgblue}0.72 $\pm$ 0.038
   & \cellcolor{kgblue}0.76 $\pm$ 0.038
   & \cellcolor{kgblue}0.75 $\pm$ 0.052
   & \cellcolor{kgblue}0.70 $\pm$ 0.041
   & \cellcolor{kgblue}0.74 $\pm$ 0.040
   & \cellcolor{kgblue}0.73 $\pm$ 0.057 \\
\bottomrule
\end{tabular}%
}
\label{tab:gpt-gemini-merged}
\end{table*}

Table \ref{tab:gpt-gemini-merged} presents the performance of all hallucination self-detection methods, with and without knowledge-graph (+KG) support, for GPT-4o and Gemini-2.5-Flash. Across datasets and metrics, a clear pattern emerges: adding knowledge graphs consistently improves accuracy, F1, and AUC-PR for every method, often by substantial margins. We organise the discussion by presenting the research questions (in bold) and the results on which we can base our conclusions.

\vspace{0.1cm}
\noindent\textbf{Does including knowledge graphs in the self-detection mechanism improve its performance?} Across both GPT-4o and Gemini-2.5-Flash, adding knowledge-graph structure consistently boosts hallucination detection, improving accuracy by up to 14\% (Self-Questioning on GPT-4o for SimpleQA) and F1 by up to 20\% (Self-Questioning on the same task). On average across all methods and datasets, knowledge graphs deliver a 6.8\% accuracy gain, a 9.3\% F1 gain, and a 7.5\% increase in AUC-PR, demonstrating broad and reliable benefit. For both models, the improvements obtained by adding knowledge graphs are statistically significant at the 95 percent level for all metrics. Across both models, KG increases AUC-PR consistently, showing that improvements hold regardless of decision threshold and reflect true calibration gains. This confirms that, as anticipated, fact-level decomposition offers a sharper and more reliable signal for hallucination detection, and that isolating and analysing structured facts instead of plain text is much more effective and interpretable for larger models.

\vspace{0.1cm}
\noindent\textbf{Which self-detection approach yields the best results?} Across all models, Self-Confidence emerges as the strongest overall method when compared across the WikiBio GPT-4o dataset, reaching the highest absolute F1 scores, with and without knowledge graphs. This suggests that directly eliciting the model’s own confidence provides a reliable and interpretable signal of factuality. In terms of F1 scores, the Self-Questioning method shows the highest improvement using knowledge graphs across both models, with a 16\% improvement for GPT-4o and 17\% for Gemini across both datasets. SelfCheckGPT has the highest accuracy and AUC-PR gains when supplemented with knowledge graphs.

\vspace{0.1cm}
\noindent\textbf{How convenient is using a multi-sample hallucination self-detection technique?} Multi-sample methods on WikiBio GPT-4o fail to outperform single-sample approaches, indicating that generating and comparing multiple responses may not justify the extra computational expense. This gap likely arises because powerful models like GPT-4o and Gemini-2.5-Flash already produce relatively stable outputs, limiting the added value of cross-sample consistency checks and making fact-level single-sample scoring a more efficient strategy.

\vspace{0.1cm}
\noindent\textbf{How can we interpret the performance improvement with KGs?} The improved performance obtained with knowledge graphs indicates that LLMs often “know” more correct individual facts than their hallucinated outputs reveal. Breaking responses into knowledge-graph triples delivers higher accuracy and F1, showing that hallucinations can be effectively exposed even without external evidence. We suggest that hallucination research should focus less on whole-text inferences and more on how models internally connect or present facts, particularly when context is missing or misleading and correct facts are combined into incorrect or ambiguous statements.

\section{\uppercase{Conclusion}}
Our study shows that the proposed knowledge graph driven procedure consistently improves hallucination self-detection across strong models such as GPT-4o and Gemini-2.5-Flash. Among the evaluated methods, Self-Confidence achieves the highest absolute F1 scores, while Self-Questioning benefits most from KG integration and SelfCheckGPT shows the largest gains in accuracy and AUC-PR. Multi-sample methods do not outperform single-sample approaches, suggesting that the computational cost of repeated generation offers limited benefit when model outputs are already stable. Fact-level analysis further indicates that hallucinations often arise from incorrect fact associations or misleading context rather than missing atomic facts, highlighting the importance of how models link and present information.

The proposed approach requires only the addition of self-constructed knowledge graphs to calibrate hallucination scores, without training or parameter tuning. However, our analysis is limited to best-performing thresholds, and score interpretation is method dependent. Experiments are restricted to English datasets and a relatively small multi-sample test set, leaving open exploration around cross-lingual generalisation and multi-sample robustness. Extending KG-driven fact analysis across languages, domains, and evaluation regimes will be critical for developing more reliable and controllable LLMs.

\bibliographystyle{apalike}
{\small
\bibliography{main}}
\end{document}